\documentclass[letterpaper]{article}
\usepackage{mine}  
\usepackage{times}
\usepackage{helvet}
\usepackage{courier}
\usepackage{times}
\usepackage{latexsym}
\usepackage{booktabs} 
\usepackage{tabularx}
\usepackage{amsmath}
\usepackage{subcaption} 
\usepackage{multirow}
\usepackage{algorithm}
\usepackage{algpseudocode}
\usepackage{xcolor} 
\usepackage{mdframed} 
\usepackage{lipsum} 
\definecolor{lightgray}{gray}{0.9}
\usepackage{listings}
\usepackage{graphicx}
\usepackage{amssymb}
\usepackage{booktabs}
\usepackage{colortbl}
\usepackage{xcolor}
\usepackage[utf8]{inputenc}
\DeclareUnicodeCharacter{0394}{$\Delta$}

\usepackage{makecell}
\usepackage{amssymb} 
\usepackage[utf8]{inputenc}
\usepackage{caption}  
\usepackage{float}
\usepackage{stfloats}
\usepackage{microtype}
\usepackage{tcolorbox}
\usepackage{natbib}
\usepackage{tikz}
\usepackage{geometry}
\newtcolorbox{mybox}{
  colback=red!10, 
  colframe=red!75!black, 
  boxrule=1pt, 
  arc=4pt, 
  boxsep=0pt, 
  left=6pt, 
  right=6pt, 
  top=6pt, 
  bottom=6pt, 
  width=\textwidth 
}

\newtcolorbox{mybox1}{
  colback=gray!10, 
  colframe=gray!75!black, 
  boxrule=1pt, 
  arc=4pt, 
  boxsep=0pt, 
  left=6pt, 
  right=6pt, 
  top=6pt, 
  bottom=6pt, 
  width=\textwidth 
}

\newtcolorbox{mybox0}{
  colback=gray!10, 
  colframe=gray!10, 
  boxrule=0pt, 
  arc=4pt, 
  boxsep=0pt, 
  left=14pt, 
  right=6pt, 
  top=1pt, 
  bottom=1pt, 
  width=\textwidth 
}

\begin{document}
%
\title{Role-RL: Online Long-Context Processing with Role Reinforcement Learning for Distinct LLMs in Their Optimal Roles}

\author {
    Lewei He\textsuperscript{\rm 1},
    Tianyu Shi\textsuperscript{\rm 2},
    Pengran Huang\textsuperscript{\rm 1},
    Bingzhi Chen\textsuperscript{\rm 1},
    Qianglong Chen\textsuperscript{\rm 3}\textsuperscript{*},
    Jiahui Pan\textsuperscript{\rm 1}\textsuperscript{*}
}
\affiliations {
    \textsuperscript{\rm 1}School of Artificial Intelligence, South China Normal University\\
    \textsuperscript{\rm 2}Transportation Research Institute, Toronto University\\
    \textsuperscript{\rm 3}College of Computer Science and Technology, Zhejiang University\\
    \{helewei, huangpengran, chenbingzhi, panjiahui\}@m.scnu.edu.cn,\\ ty.shi@mail.utoronto.ca, chenqianglong@zju.edu
}

\maketitle

\begin{abstract}
\begin{quote}


Large language models (LLMs) with long-context processing are still challenging because of their implementation complexity, training efficiency and data sparsity. To address this issue, a new paradigm named Online Long-context Processing (OLP) is proposed when we process a document of unlimited length, which typically occurs in the information reception and organization of diverse streaming media such as automated news reporting, live e-commerce, and viral short videos. Moreover, a dilemma was often encountered when we tried to select the most suitable LLM from a large number of LLMs amidst explosive growth aiming for outstanding performance, affordable prices, and short response delays. In view of this, we also develop Role Reinforcement Learning (Role-RL) to automatically deploy different LLMs in their respective roles within the OLP pipeline according to their actual performance. Extensive experiments are conducted on our OLP-MINI dataset and it is found that OLP with Role-RL framework achieves OLP benchmark with an average recall rate of 93.2\% and the LLM cost saved by 79.4\%. The code and dataset are publicly available at: https://anonymous.4open.science/r/Role-RL.
\end{quote}
\end{abstract}

\section{Introduction}
In the academic and technological spheres, large language models (LLMs) have emerged as pivotal tools for improving production efficiency and advancing our understanding of human language. 
These models leverage deep learning techniques, particularly the power of transformers, to capture the intricacies of syntax, semantics, and context within both oral and written communications. 

As computational linguistics makes a significant leap forward, a growing number of LLMs are quickly coming to the forefront. 
However, the associated problems also become more pronounced, that is, choosing the most suitable LLM for a given task. Studies have shown that different LLMs have different strengths, weaknesses, and even personalities due to differences in training datasets \citep{pe2}. For example, generative model T5 has superior performance to discriminative model Bert on conversational QA tasks \citep{ab1}. For zero-shot coding assignments, InCoder-1b scores twice as much as Code Llama-7b, but the situation reverses in 1-shot tasks, with Code Llama-7b scoring considerably higher \citep{ab2}. In terms of comparative reasoning, multimodal LLM LLaVA-1.6 has better performance in spatiality comparison tasks whereas GPT-4V is stronger in temporality comparison tasks \citep{ab3}. In terms of personality, GPT-Neo is more extroverted than GPT-3.5, while GPT-3.5 exhibits a higher degree of agreeableness \citep{pe1}\citep{pe5}. It was also found that ChatGPT embodies the ENTJ personality type in the MBTI framework, characterized by self-confidence, decisiveness, and natural leadership abilities, whereas OpenLlama7b aligns with the INFJ type, known for its deep insight into people and a strong adherence to personal values \citep{pe4}. Hence, it is essential to carefully choose the most appropriate LLMs on the basis of their unique characteristics and the role requirements they are intended to fulfill.

\begin{figure}
    \centering
    \includegraphics[width=0.95\linewidth]{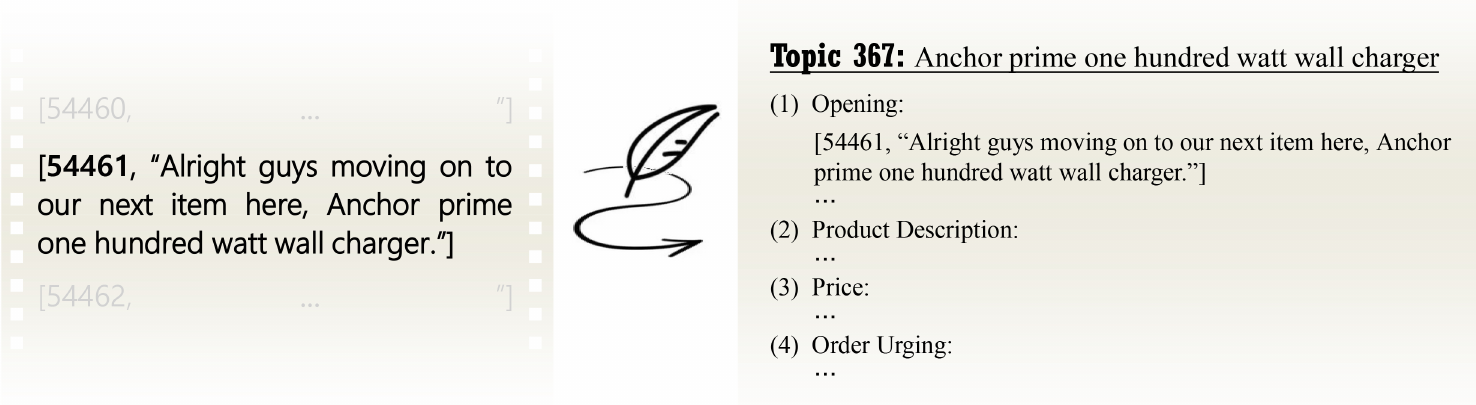}
    \caption{Illustration of Online Long-context Processing (OLP) problem.}
    \label{fig0}
\end{figure}

Another issue regarding LLM is the ability to process long contexts, especially in the scenario of streaming media transcripts of unlimited length, as shown in Figure \ref{fig0}. Streaming media refers to an emerging form of media content that is continuously delivered without requiring a complete download before playback, which is becoming increasingly relied upon by the public \citep{md2}. A survey of young people between 16 and 24 years of age revealed that 55.1\% of the respondents spent 2--3 hours, 16.3\% spent 3--4 hours, and 12.2\% spent more than 4 hours on streaming media every day \citep{md}. This means that an end user consumes approximately 27k words per day if a speed of 2.5 words per second is assumed \citep{spk} for language-intensive content such as live e-commerce, which is comparable to a novella with 20k--40k words. From the perspective of end users, it is helpful if an excerpt of transcripts is generated in real-time to help them trace back their desired products and key selling points during the live-stream shopping. From the perspective of enterprise users, it is indispensable to analyze their many competitors instantly and learn from the top streamers in specific aspects, such as how order urging is performed for each product in a natural but effective way.

\begin{figure}[t]
    \centering
    \includegraphics[width=\linewidth]{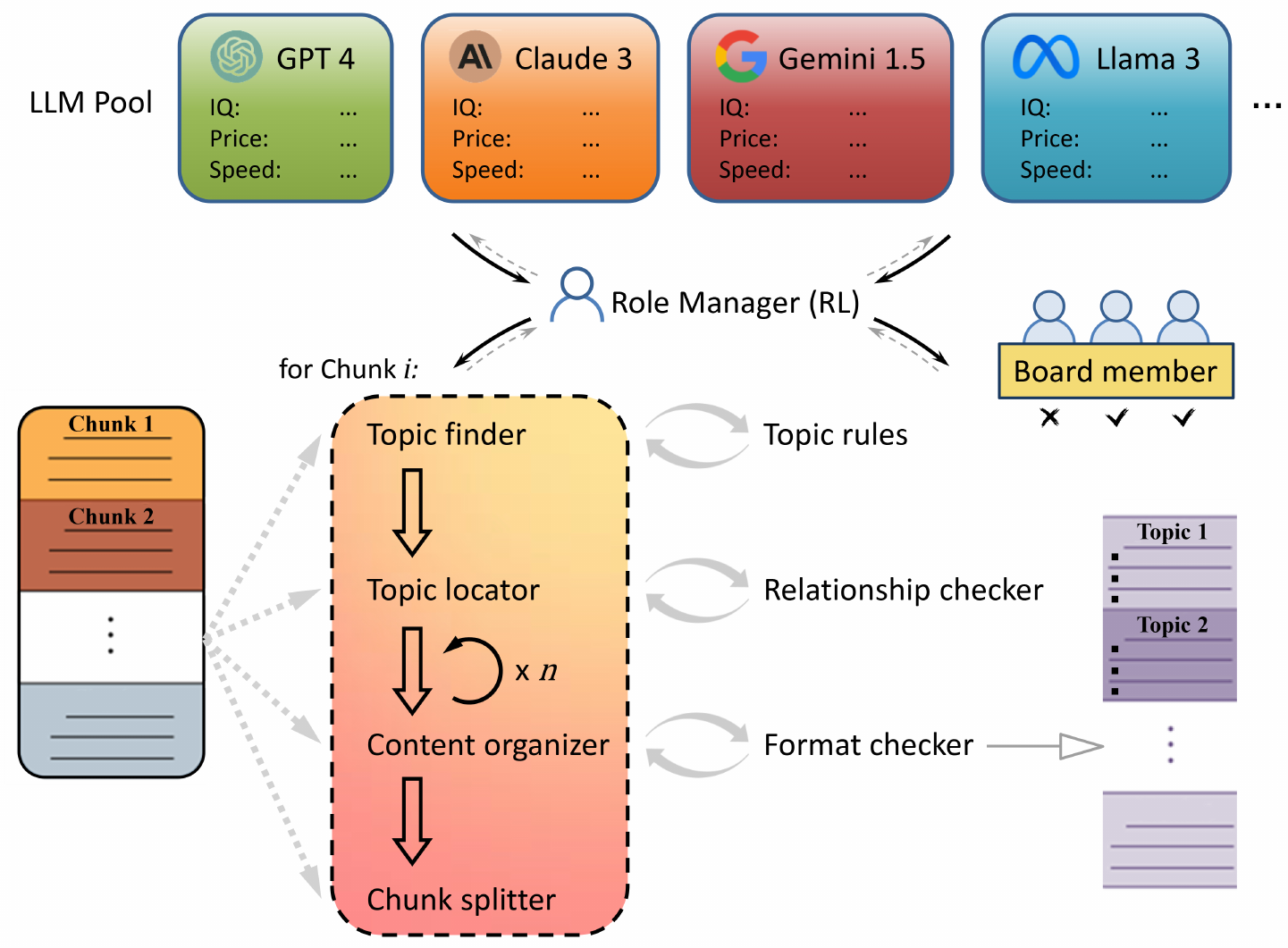}
    \caption{Architecture of the proposed Online Long-context Processing (OLP) pipeline in Role Reinforcement Learning (Role-RL) framework. The OLP pipeline consists of six well-defined roles that collaborate effectively to extract useful information from the context with unlimited length and restructure them into topics with supportive aspects. The Role-RL framework is composed of an LLM pool, an LLM advisory board, and a Role Manager driven by reinforcement learning to place the LLMs in different roles according to their actual performances and costs.
}
    \label{fig1}
\end{figure}

To address these issues, we propose an Online Long-context Processing (OLP) pipeline with a Role Reinforcement Learning (Role-RL) framework, as shown in Figure \ref{fig1}. The OLP pipeline consists of Topic finder, Topic locator, Relationship checker, Content organizer, Format checker and Chunk splitter, with all or part of them driven by LLMs. They collaborate to process the transcripts, organizing them into various topics and categorizing the original passages according to different aspects in an online manner while the transcripts are coming in. In the scenario of live e-commerce, a topic refers to a product for sale and the aspects may include “Opening”, “Product description”, “Price” and “Order urging”, while in the scenario of automated news reporting, a topic refers to a news event and the aspects may include “Facts”, “Opinions”, “Assumptions” and “Future plans”. Additionally, in the Role-RL framework, an advisory board made up of LLMs observes each LLM in the OLP pipeline and gives judgments on their performance. A role manager empowered by reinforcement learning calculates the reward earned by each LLM on the basis of their output correctness, API cost and response delay, and then places the right LLMs in the right roles to improve the performance of the whole system. Based on the LLM rankings implied in the process of reinforcement learning, the board members are elected and updated regularly to increase the credibility of their judgments.

The main contributions of our work are as follows:
\begin{itemize}
    \item We introduce Role-RL, a reinforcement learning framework to deploy distinct LLMs in the optimal roles according to their actual performances.
    \item We propose a new pipeline named OLP in the realm of long-context processing and contribute to the corresponding dataset OLP-MINI.
    \item Our Role-RL framework achieves an OLP benchmark with an average recall rate of 93.2\% and LLM cost saving of 79.4\%, accompanied by the OLP pipeline which increases the recall rate by 53.6 percentage points compared to non-OLP techniques.
\end{itemize}

\section{Related work}

\subsection{Long-context processing}
LLM on long-context tasks is a popular research topic and many techniques were developed to improve the long-context processing ability.
Infinite-LLM \citep{long1} manages dynamic context lengths efficiently through a distributed attention mechanism called DistAttention and a pooled GPU memory strategy, enabling support for extensive context lengths up to 2000K tokens and demonstrating a 1.35--3.4x throughput improvement.
MEGALODON \citep{long2} features timestep normalization and normalized attention for efficient long context handling, outperforming Transformers in efficiency and accuracy especially in pretraining and downstream tasks.
LongRoPE \citep{long3} expands LLM context to 2048k tokens with an efficient search algorithm that exploits non-uniformities in positional interpolation and a progressive extension strategy.
SelfExtend \citep{long4} enhances LLMs' long context handling via a bi-level attention mechanism which maps large relative positions to known ones with a simple floor division operation.

Meanwhile, long-context processing by agent cooperation is another direction. CoA \citep{COA} improves LLMs in long-context tasks by aggregating information and reasoning across models with enhanced performance in QA, summarization, and code completion.
GraphReader \citep{graphreader} structures long texts into a graph for autonomous exploration, outperforming other models in long-context QA and showing robustness in handling very long texts.
LONGAGENT \citep{longagent} scales LLMs to manage up to 128K tokens by dividing contexts, assigning agents, and resolving conflicts, with an average improvement of 19.53\% in single-document.
PEARL \citep{pearl} enhances reasoning over long documents by segmenting the process into action mining, plan formulation, and execution, showing effectiveness on a subset of the QuALITY dataset.

\subsection{Novel agent frameworks}
Beyond agent frameworks tailored for the long-context task, numerous insights and methodologies can be gleaned from a variety of multipurpose frameworks, for example, to elect an LLM leader \citep{elect} and add agents into the system dynamically and automatically \citep{harness}. 
In addition, AgentCoord \citep{agentcoord} develops a visual interface for multi-agent coordination strategy design using a three-stage generation method. 
PRD \citep{PRD} utilizes Peer Rank and Discussion with varying judge weights for LLM assessment to reduce biases and align with human judgments.
MAD \citep{MAD} addresses Degeneration-of-Thought with affirmative and negative debaters in self-reflection with enhanced performance in complex reasoning tasks.
DyLAN \citep{DyLAN} features multi-round interactions and an agent selection algorithm based on the Agent Importance Score, with improved accuracy and efficiency in reasoning and code generation. 
RL-GPT \citep{RL-GPT} introduces a two-level hierarchical framework to enhance agent performance in complex, open-world environments by leveraging the strengths of both high-level planning and low-level action.

\section{Problem formulation}
A transcript of unlimited length allows for an unlimited number of potential chunks, represented by $\{c_1, c_2, ...\}$. In the chunk $i$, there are several topics $\{t_1, t_2, ..., t_m\}$. For the topic $j$, we are concerned about $n$ aspects $\{a_{j,1}, a_{j,2}, ..., a_{j,n}\}$ which are predefined and usually kept the same for all topics. If we assume the first passage related to topic $j$ is $p_{j,1}$ and the last is $p_{j,-1}$, for topic $j$ we have a set of passages $p_j = \{p_{j,1}, ..., p_{j,-1}\}$. In order to organize $p_j$ across different aspects, we aim to find $p_j \cap a_{j,k}$ with $k = 1 \sim n$ for the $n$ aspects of the topic $j$. Similarly, the process also applies to each and every other topic within a given transcript.

\section{Method}

\subsection{Online Long-context Processing (OLP)}
Streaming media has become more and more important in our daily lives. However, it is excessively laborious for humans to distill useful information from streaming media, and this task also does not fit well into conventional long-context processing due to its unlimited length of the context and the demand for real-time output with online processing ability. In order to address this issue and organize the transcripts of streaming media according to the required specifications with minimal repetitions, losses and hallucinations, we propose Online Long-context Processing (OLP) pipeline. As shown in Figure \ref{fig1}, OLP splits the streaming long context into chunks automatically according to its length and semantic information, and processes each chunk into structured text with core topics and related aspects of supportive information listed below. It consists of Topic finder, Topic locator, Relationship checker, Content organizer, Format checker, and Chunk splitter, which cooperate closely to classify each passage of the streaming long context into the aspects that we focus on. 

Topic finder recognizes the topics that we are interested in, for instance products for sale, news events, and short video storylines. Topic locator locates the content correlated to each topic, usually by the passage index that is distributed for each passage as its unique ID. Relationship checker examines the topics that are recognized in the first step based on their semantic and location information, and decides whether to delete a topic or merge two relevant topics together. With the confirmed topics and locations, Content organizer retrieves the passages relevant to each topic except the last topic in a chunk, and reorganizes the passages into a structured layout topic by topic with supportive passages below each key aspect. Finally, Chunk splitter divides the current chunk between the last topic and other topics, and passes the passages of the last topic to the next chunk in order to avoid the scenario where a single topic is divided into two chunks with two different titles allocated. As such, OLP guarantees that the topics are recognized properly and the passages in each aspect of a topic are excerpted exhaustively.

\subsection{Role Reinforcement Learning (Role-RL)}
A dilemma was often encountered when we sought the optimal LLM from already numerous and still growing LLMs with outstanding performance, affordable price and short response delay. Lots of experimental works were required to test the LLMs one by one, especially in scenarios like OLP where the roles are highly specialized with very distinct difficulty levels and response delay requests. Worse still, these labor works are permanent since the role requirements may evolve and there are always new LLMs coming out to be compared with the LLMs already in position. Therefore, Role-RL is developed by this work to select the LLMs automatically and place them into the most suitable roles with optimized overall performance of an LLM network.

\begin{figure}[htbp]
    \centering
    \includegraphics[width=0.7\linewidth]{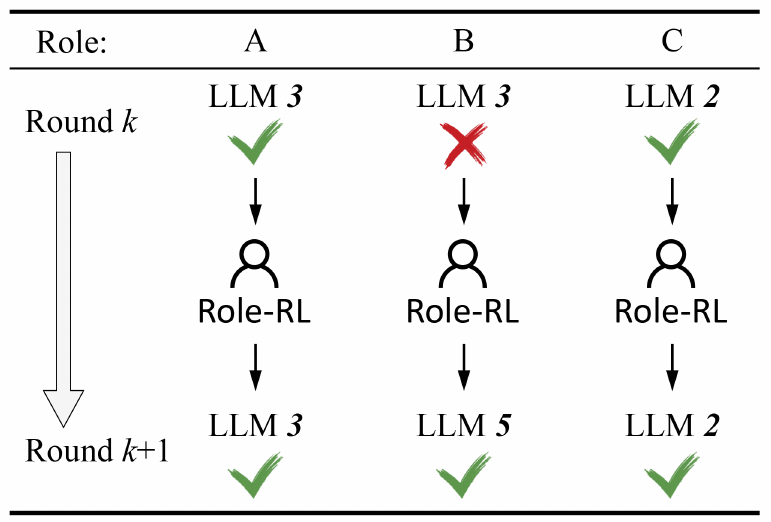}
    \caption{Illustration of Role-RL functionality.
}
    \label{roleRL}
\end{figure}

\subsubsection{Q-learning and reward.}
Ideally, Role-RL should maintain an LLM that successfully completed the previous round, but assign a more potent LLM if the previous attempt was unsuccessful, as shown in Figure \ref{roleRL}. Therefore in the reinforcement learning, the possible actions $\mathcal{A}$ are the LLMs in the pool to select from, and the state $S_t$ is the LLM used in the previous round with a success or failure resulted. Given the discrete actions and states in a limited number, Q-learning is adopted as the base of Role-RL due to its robustness and efficiency. The reward function in Role-RL can be tailored to specific objectives and conditions, but typically it is expressed by
\begin{equation}
R = v - k_1c - k_2t
\end{equation}
where $v$ stands for the reward due to the accuracy of the answer which is judged by the board members, $c$ for the LLM cost, $t$ for the response delay, and $k_1$ and $k_2$ for the coefficients of $c$ and $t$. In this study, an LLM is deemed successful in a role when it receives a positive reward, and unsuccessful with a negative reward.

\begin{figure}[tp]
    \centering
    \includegraphics[width=0.6\linewidth]{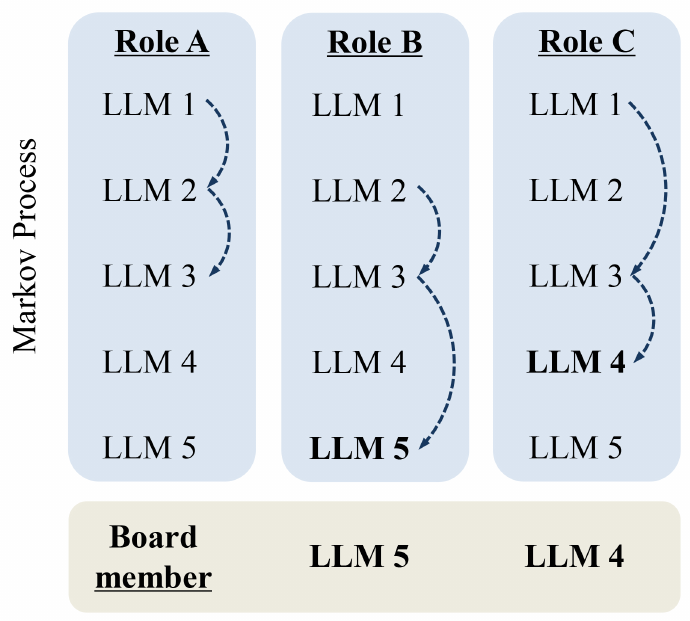}
    \caption{Election of the board members.
}
    \label{board}
\end{figure}

\subsubsection{Update of board member and LLM pool.}
Since the accuracy of an LLM answer is judged by the board members, they should be elected from the LLMs with relatively high inference ability to increase the confidence level of their judgments. Moreover, the election of board members should also be automatic without generating too much burden on the system. In order to achieve this, Markov chains are extracted from the failure states in the Q-table, since only stronger LLMs receive positive rewards when an LLM fails, which in turn implies an inherent ranking order of those LLMs. Taking Figure \ref{board} as an example, the Markov chain in Role A ends at LLM 3, indicating that LLM 3 is enough to solve all the tasks in Role A. However in Role B and Role C, the Markov chains go through LLM 3 and end at LLM 5 and LLM 4 respectively, which implies that LLM 5 and LLM 4 are stronger than LLM 3 and thus selected as the board members to judge the answers of all the LLMs. Moreover, the board members should be updated according to the current Markov chains and Q-tables, and the update of board members should be smooth to reduce any fluctuations imposed on the system. Therefore, a maximum change limit denoted by $\Delta w$ is applied to the voting weight of each board member, and the final reward is calculated by a weighted sum of the rewards from all board members. That is, an LLM is excluded from the advisory board if its weight is equal to 0. Furthermore, with the evolution of LLMs, the LLM pool should be continually updated with new models that offer enhanced capabilities at a lower cost, and the LLM pool can either be expanded or remain the same size by replacing the LLM with the lowest success rate.

\subsubsection{Greedy-update and cross-update strategies.}

\begin{figure}[tp]
    \centering
    \includegraphics[width=\linewidth]{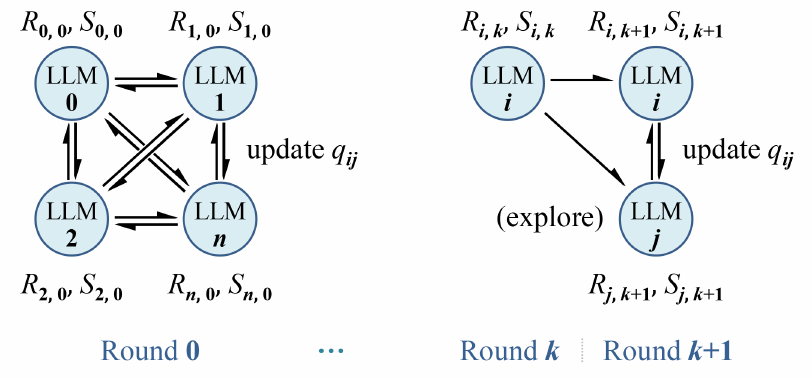}
    \caption{Schematics of greedy-update (left) and cross-update (right) strategies.
}
    \label{update}
\end{figure}

As shown in Figure \ref{update}, a greedy-update strategy is developed to use the data more effectively and avoid a cold start during the deployment of Role-RL. In Round 0, every LLM in the pool takes the role in turn and obtains respective reward $R_{i,0}$ and state $S_{i,0}$ where $i$ ranges from 0 to the maximum LLM index $n$. 
Then one LLM serves as the previous LLM (e.g. LLM 0) 
and all other LLMs (e.g. LLM 1 $\sim$ LLM $n$) serve as the current LLM in turn
to update the Q-table with the previous state (i.e. $S_{0,0}$) and current rewards (i.e. $R_{1,0} \sim  R_{n,0}$). 
Every other LLM also serves as the previous LLM in turn until half of the Q-table is filled, since the outcome of each LLM can be either success or failure and the q-values of the other states are left as 0. 
After Q-table update, only the LLM answer with the highest reward is returned and this LLM is stored in the LLM history. Similarly, in the cross-update strategy, when the current LLM (e.g. LLM $j$) is different from the previous LLM (e.g. LLM $i$) due to either exploration or greedy choice of the best action, 
both LLMs are tested and the q-values are updated with four state-reward pairs, namely $S_{i,k}$-$R_{i,k+1}$, $S_{i,k}$-$R_{j,k+1}$, $S_{i,k+1}$-$R_{j,k+1}$, and $S_{j,k+1}$-$R_{i,k+1}$. 
This cross-update strategy with both the previous and current selections of LLM also improves the system robustness and guarantees that in each round any new choice of LLM would not be worse than keeping the LLM unchanged.

\begin{figure}[bp]
    \centering
    \includegraphics[width=0.89\linewidth]{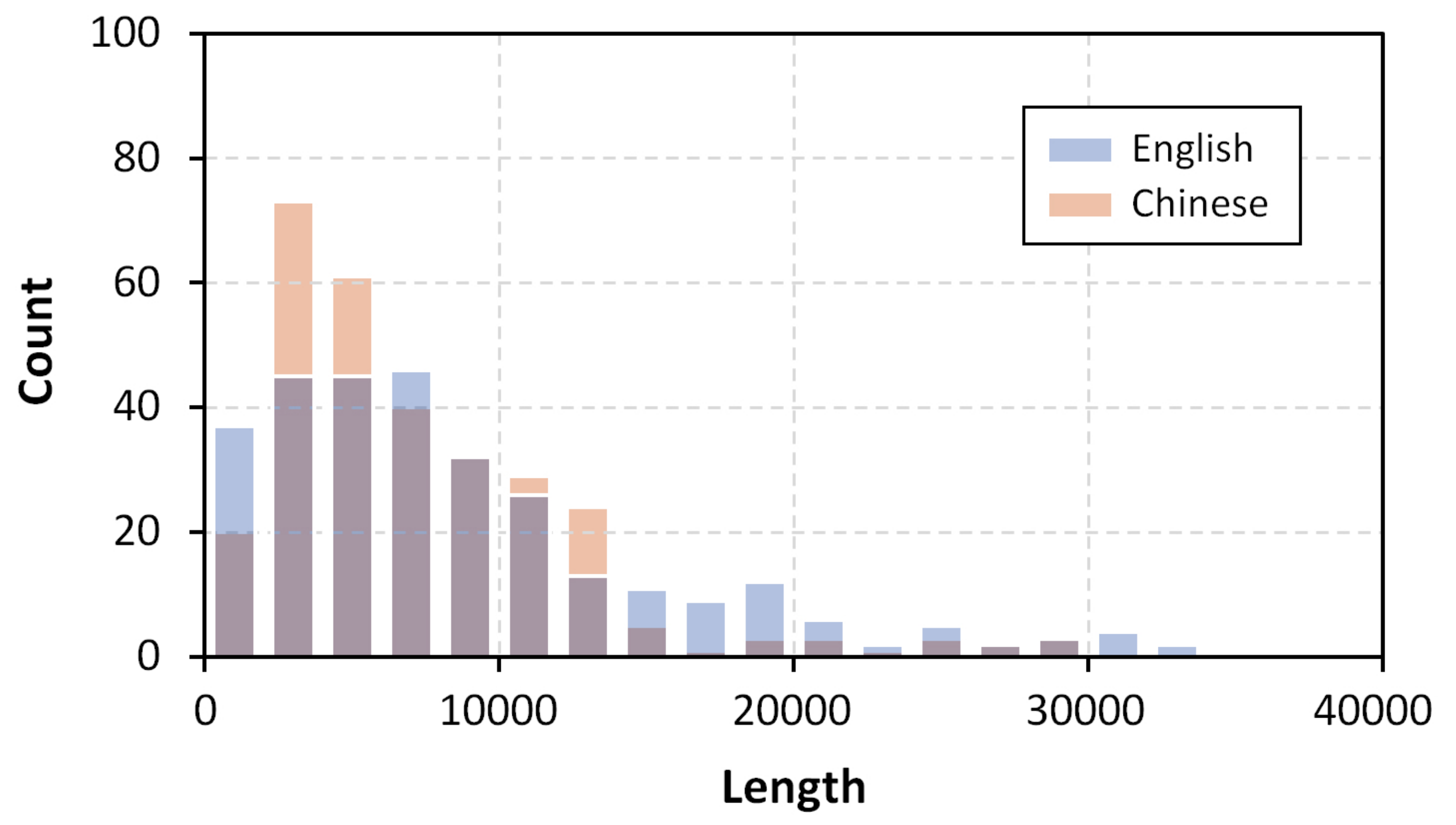}
    \caption{Length statistics of English and Chinese transcripts in the OLP-MINI dataset. 
}
    \label{dataset}
\end{figure}

\subsection{Datasets}
The dataset for OLP can be transformed from any long-context datasets by adding a unique index at the front of each passage for topic locating. We also contribute our dataset OLP-MINI which consists of 300 English live-commerce transcripts from TikTok and 300 Chinese live-commerce transcripts from Alibaba. The statistics of transcript length is shown in Figure \ref{dataset}, with an average length of 8758 words for English transcripts and 7304 characters for Chinese transcripts.

\begin{figure*}[tp]
    \centering
    \includegraphics[width=0.73\linewidth]{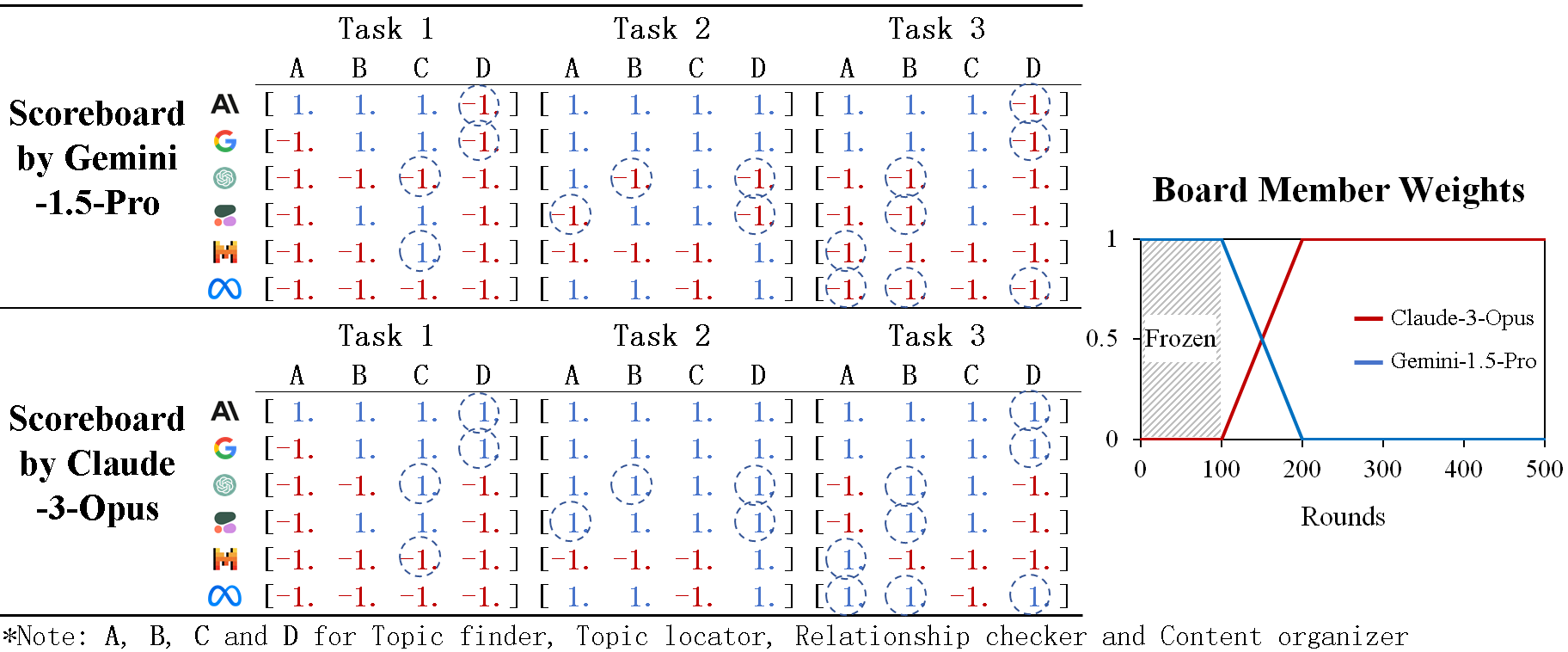}
    \caption{The scoreboards judged by Gemini-1.5-Pro and Claude-3-Opus (left) and the weights of board members in the first 500 rounds (right). 
}
    \label{result1}
\end{figure*}

\section{Experiment}

\subsection{Settings}
The experiments are conducted on the OLP-MINI dataset with four aspects listed for each topic, namely “Opening”, “Product description”, “Price” and “Order urging”. Topic finder, Topic locator, Relationship checker and Content organizer are empowered by LLMs, whereas Format checker and Chunk splitter are written in hard code. Learning rate $\alpha=0.1$ and exploration rate $\epsilon=0.03$ are adopted with 50k rounds of reinforcement learning on three tasks, namely Task 1 with 4 topics and about 676.3 characters per topic, Task 2 with 2 topics and about 139.0 characters per topic, and Task 3 with 8 topics and about 253.5 characters per topic. In each round, the task in the previous round continues unless random task replacing with a probability $\omega = 0.03$. Attenuation factor $\gamma=0$ is adopted since the LLM with the highest reward is desired in each round. In the reward function, $v=10$ for the correct answer and $v=-100$ for the wrong answer are given to Topic finder and Relationship checker, while IoU and similarity are calculated for Topic locator and Content organizer to map $v$ between 10 and -100. The response delay $t = 1$ s remains for every LLM to avoid any unfairness due to the network with the cost and time coefficients $k_1 = 1$ and $k_2 = 0.1$. Six LLMs are selected in the initial LLM pool, namely Llama3-8b, Mixtral-8x7b, Command-r, GPT-4o, Gemini-1.5-Pro, and Claude-3-Opus, whereas GPT-4o-mini is added in the last 500 rounds to replace the LLM with the lowest success rate in each role. If there are two LLMs with the lowest success rate, the more expensive one will be replaced. The price of each LLM is shown in Table \ref{tab_cost}, while the input and output token consumptions are fixed to 25k for a close comparison between LLMs in different roles and tasks.

\begin{table*}[bp]
\centering
\resizebox{0.85\textwidth}{!}{
\begin{tabular}{lccccccc}
\toprule
(\$ / 1M tokens) & \textbf{Llama3-8b} & \textbf{GPT-4o-mini*} & \textbf{Mixtral-8x7b} & \textbf{Command-r} & \textbf{GPT-4o} & \textbf{Gemini-1.5-Pro} & \textbf{Claude-3-Opus} \\ \midrule
Input & 0.08 & 0.15 & 0.7 & 0.5 & 5 & 7 & 15 \\
Output & 0.08 & 0.6 & 0.7 & 1.5 & 15 & 21 & 75 \\ \bottomrule
\multicolumn{8}{l}{* added in the last 500 rounds}
\end{tabular}
}
\caption{The cost of LLMs that are involved in the experiment.}
\label{tab_cost}
\end{table*}

In this work, the board members judge the output of LLMs in every round, whereas in practice it is feasible to judge the output only on a regular basis and when the LLM choice changes. When the board members do not fully agree with an answer, they revise it to the correct form and then pass it on to subsequent roles. Gemini-1.5-Pro is nominated as the initial board member with maximum weight step $\Delta w = 0.01$. The weights are frozen in the first 100 rounds until the Q-tables are well updated, and remain unchanged if a loop is formed in the chain. Notably, the four Q-tables by Topic finder, Topic locator, Relationship checker and Content organizer are added together as a representative Q-table, and then combined with a cost table to eliminate the impact of the LLM price and reflect the real rankings between LLMs in terms of their context comprehension ability.

\subsection{Main Results}

The scoreboards for each LLM, each role, and each task judged by Gemini-1.5-Pro and Claude-3-Opus are shown in Figure \ref{result1}, with a success noted by a blue “1”, a failure noted by a red “-1”, and the different scores between two scoreboards marked by dotted circles. Task 2 is relatively simpler with more correct answers, and the first two LLMs, especially Claude-3-Opus, exhibit a higher success rate than the other LLMs. By comparing the two scoreboards and the specific judgments from the two board members, Gemini-1.5-Pro sometimes marks an answer as wrong and then corrects the original answer with exactly the same one, which indicates that it is not quite qualified as a board member to judge the answers objectively. Owing to a higher success rate, the Markov chain extracted from the Q-tables ends at Claude-3-Opus; thus, its weight on the board starts increasing after the first one hundred rounds with frozen weights and becomes the sole board member by replacing Gemini-1.5-Pro in the 200\textsuperscript{th} round.

\begin{figure}[tbp]
    \centering
    \includegraphics[width=\linewidth]{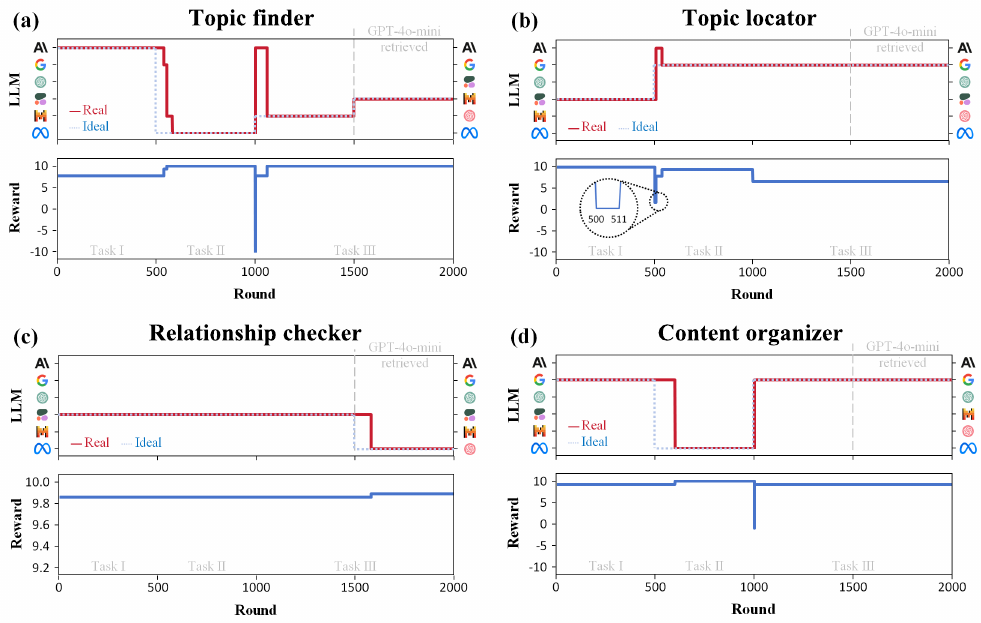}
    \caption{LLM and reward histories for (a) Topic finder, (b) Topic locator, (c) Relationship checker and (d) Content organizer.
}
    \label{result2}
\end{figure}

The choices of LLM and the corresponding rewards for the four roles, namely Topic finder, Topic locator, Relationship checker, and Content organizer are demonstrated on Task 1 for 500 rounds followed by Task 2 for 500 rounds and Task 3 for 1000 rounds. As shown in Figure \ref{result2}(a) for the Topic finder, Claude-3-Opus is handling Task 1 during the first 500 rounds although it is the most expensive LLM, probably because it is challenging to find the topics correctly with longer context for each topic. During the second 500 rounds, the LLM shifts from Claude-3-Opus to Llama3-8b due to exploration for a higher reward, since Task 2 is easier with fewer topics and shorter context. At the 1000\textsuperscript{th} round, Task 2 shifts to Task 3 which is beyond the capability of Llama3-8b, and thus Llama3-8b gives a wrong answer with negative reward and “LLM 0 failed” state is entered. 
At the 1001\textsuperscript{st} round, Role-RL assigns a new LLM with the highest q-value in the “LLM 0 failed” state, which is Claude-3-Opus in this case. After several rounds, the LLM again shifts from Claude-3-Opus to Mixtral-8x7b due to the exploration for a higher reward. In the last 500 rounds, GPT-4o-mini is added to the pool with GPT-4o replaced given its relatively poor performance in this role.

As shown in Figure \ref{result2}(b) for the Topic locator, Command-r is handling Task 1 during the first 500 rounds. When the task shifts from Task 1 to Task 2, the reward drops to a small positive value. It is still considered a success so the state is “LLM 2 succeeded” and Role-RL tends to select the same LLM in the next round since this choice used to give a high reward. However, Role-RL with a learning rate $\alpha = 0.1$ recognizes that this selection is not as favorable as desired since the low reward keeps reducing the q-value of Command-r, and Claude-3-Opus becomes the new choice with the highest q-value after about 10 rounds at the 512\textsuperscript{th} round. After a while, the LLM shifts from Claude-3-Opus to Gemini-1.5-Pro through exploration and Gemini-1.5-Pro continues to serve until the end due to the highest reward even though there is a drop at the beginning of Task 3, implying that accurate topic locating is more challenging with numerous topics. In the last 500 rounds, Mixtral-8x7b is replaced by GPT-4o-mini. Notably, Tasks 2 and 3 with fewer characters per topic are relatively easier for Topic finder but harder for Topic locator. This observation is probably because topic locating with shorter context is more stringent since a slight deviation of the starting and ending passages leads to a significant difference in IoU, which is desirable since the value of each passage is generally higher in a shorter context than that in a longer context.

As shown in Figure \ref{result2}(c) for the Relationship checker, Command-r handles all the three tasks from Task 1 to Task 3, since this role is not sensitive to the length of the context by just comparing the titles of the topics. In the last 500 rounds however, Llama3-8b is replaced by GPT-4o-mini, and GPT-4o-mini starts to take the role from Command-r due to its comparable performance at a lower cost. 

As shown in Figure \ref{result2}(d) for the Content organizer, Gemini-1.5-Pro handles Task 1 during the first 500 rounds, while the LLM for Task 2 shifts from Gemini-1.5-Pro to Llama3-8b probably because the difficulty is reduced with the fewest characters per topic. At the 1000\textsuperscript{th} round when Task 3 begins, Llama3-8b fails with a negative reward and the state becomes “LLM 0 failed” again. Therefore at the 1001\textsuperscript{st} round, Role-RL looks for the LLM with the highest q-value subject to the current state from the Q-table of Content organizer, and thus the LLM bounces back instantly from Llama3-8b to Gemini-1.5-Pro. In the last 500 rounds, Command-r is replaced by GPT-4o-mini.

The final output documents subject to different board members namely Gemini-1.5-Pro and Claude-3-Opus are compared with the right answer, and the recall rates are calculated by the passage similarity under each topic. As shown in Table \ref{tab_main}, a higher average recall rate of 0.932 with a smaller variance is achieved on the three tasks when the advisory board is managed by Claude-3-Opus, compared with a lower average recall rate of 0.873 and a larger variance when the board is managed by Gemini-1.5-Pro, revealing the effectiveness of Role-RL with the board member update mechanism.

\begin{table}[htp]
\centering
\resizebox{0.9\linewidth}{!}{
\begin{tabular}{lcccc}
\toprule
 & Task 1 & Task 2 & Task 3 & Average \\ \midrule
Total length & 2705 & 278 & 2028 & - \\
Topic quantity & 4 & 2 & 8 & - \\
Length per topic & 676.3 & 139.0 & 253.5 & - \\
\begin{tabular}[c]{@{}l@{}}Recall rate\\ (Gemini as board member)\end{tabular} & 0.817 & \textbf{0.998} & 0.804 & 0.873 \\
\begin{tabular}[c]{@{}l@{}}Recall rate \\ (Claude as board member)\end{tabular} & \textbf{0.919} & 0.927 & \textbf{0.949} & \textbf{0.932} \\ \bottomrule
\end{tabular}
}
\caption{Specifics of the three tasks on OLP-MINI and comparison of the recall rates between different board members.}
\label{tab_main}
\end{table}

\subsection{Ablation Study}

\begin{figure}[bp]
    \centering
    \includegraphics[width=\linewidth]{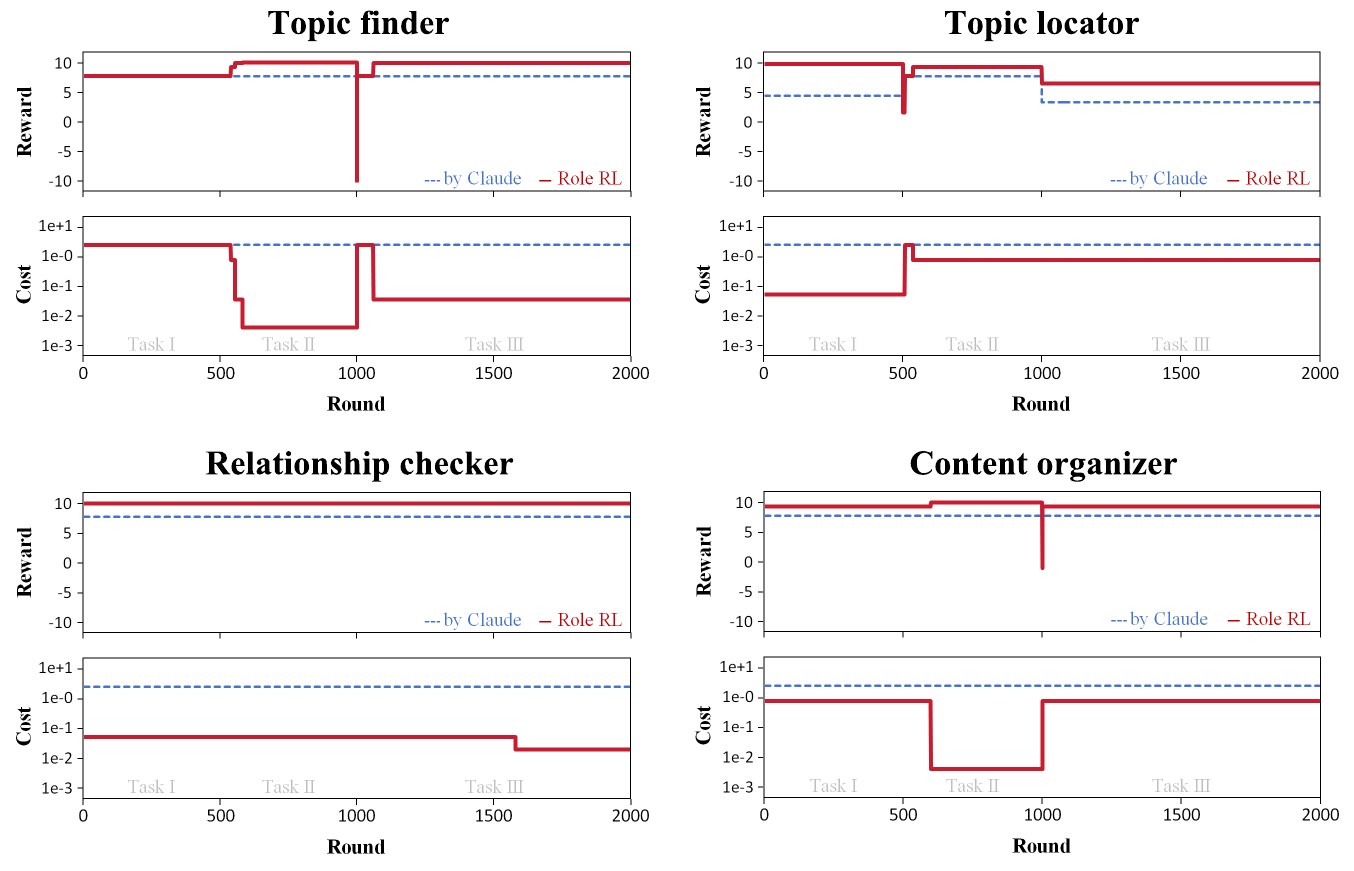}
    \caption{Reward and cost histories generated by Role-RL compared with those by solely Claude-3-Opus.
}
    \label{ablation1}
\end{figure}

To further demonstrate the effectiveness of our framework, the rewards and costs by Role-RL in Topic finder, Topic locator, Relationship checker, and Content organizer are plotted in Figure \ref{ablation1} as opposed to those solely by the board member Claude-3-Opus. For Topic finder, the cost by Role-RL drops automatically and drastically from 2.25 \$ to 0.004 \$ and then stables at 0.035 \$ depending on the difficulty of the task, which is in significant contrast to that by Claude-3-Opus consistently at 2.25 \$. For Topic locator, the cost by Role-RL starts from 0.05 \$ and ends at 0.035 \$. For Relationship checker, it remains at 0.05 \$ and drops to 0.019 \$ due to the retrieval of GPT-4o-mini. For Content organizer, it starts from 0.7 \$, then decreases to 0.004 \$ and finally rises back to 0.7 \$. Similarly, the reward by Role-RL also keeps not lower than that by Claude-3-Opus, except at a few moments when the task difficulty rises beyond those in a series of preceding rounds depending on the exploring rate $\epsilon$. By averaging the results over two thousand rounds, Role-RL significantly reduces the total LLM cost by 79.4\% and increases the total reward by 31.5\%. Moreover, a significant amount of effort is saved in updating the LLMs currently in use since the LLM repository is expandable to include new models as they are released.

\begin{figure}[htbp]
    \centering
    \includegraphics[width=\linewidth]{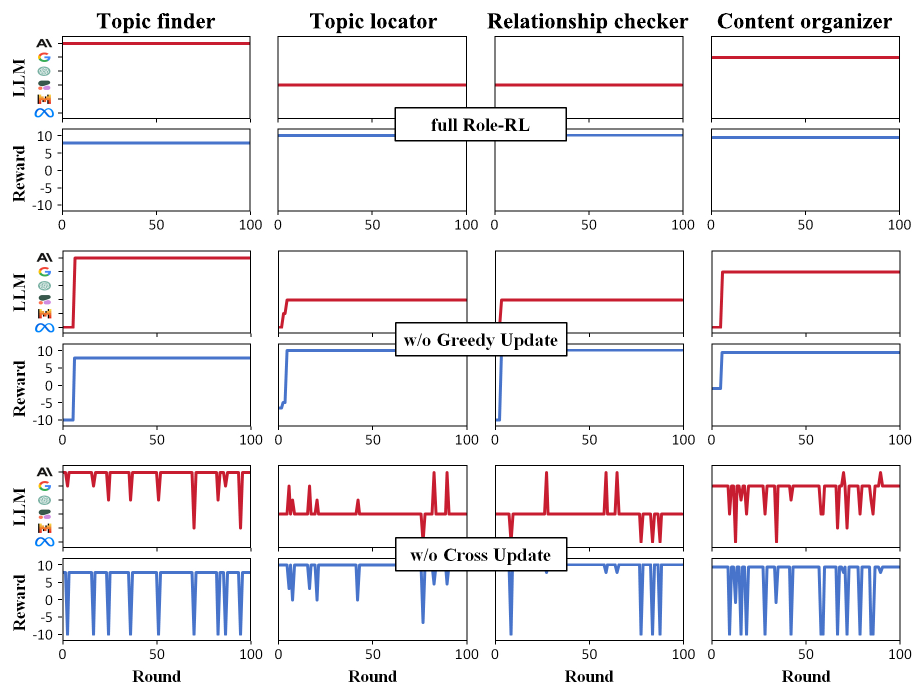}
    \caption{Reward and cost histories generated by Role-RL framework compared with the methods lacking greedy or cross updates.
}
    \label{ablation2}
\end{figure}

To illustrate the effects of greedy-update and cross-update strategies, the LLM and reward for the first 100 rounds on Task 1 are charted in Figure \ref{ablation2}, with Role-RL in contrast to the methods lacking greedy or cross updates and with the exploring rate $\epsilon$ increased from 0.03 to 0.1 for a clearer illustration. It is found that in Role-RL, the optimal LLMs are adopted from Round 0 and maintained stably throughout the process. However, LLMs and the corresponding rewards in the approach without greedy update experience a trial-and-error phase at the beginning, since the Q-tables are abundant in zeros and thus lack necessary information for Role-RL to select an LLM with positive reward. Meanwhile, LLMs in the approach without cross update easily deviate into one that is not suitable for the task, with the LLMs and rewards subjected to a series of ups and downs which is not desired in any practical applications.

\begin{table*}[htbp]
\centering
\resizebox{0.7\linewidth}{!}{
\begin{tabular}{lcccccc}
\toprule
 & MultiNews\underline{ }10 & MultiNews\underline{ }20 & MultiNews\underline{ }30 & LSHT\underline{ }10 & LSHT\underline{ }20 & LSHT\underline{ }30 \\ \midrule
Length per subject & 578.2 & 558.2 & 552.5 & 501.6 & 556.9 & 633.3 \\
Direct output & 35.9\% & 33.9\% & 25.6\% & 68.7\% & 37.3\% & 28.5\% \\
COT \citep{COT} & 44.4\% & 38.5\% & 29.3\% & 70.1\% & 39.1\% & 30.2\% \\
Reflexion \citep{reflexion} & 42.2\% & 31.7\% & 31.2\% & 71.9\% & 34.9\% & 33.2\% \\
COA \citep{COA} & 15.9\% & 9.2\% & 5.0\% & 20.7\% & 11.4\% & 6.3\% \\
LongAgent \citep{longagent} & 40.2\% & 38.6\% & 26.1\% & 69.2\% & 52.4\% & 32.4\% \\
OLP pipeline (Ours) & \textbf{85.3\%} & \textbf{87.2\%} & \textbf{84.9\%} & \textbf{93.6\%} & \textbf{89.5\%} & \textbf{92.1\%} \\ \bottomrule
\end{tabular}
}
\caption{Comparison of the recall rates with and without OLP on LongBench MultiNews and LSHT datasets \citep{longbench}.}
\label{tab_ablation}
\end{table*}

Finally, to verify the effectiveness and universality of the OLP pipeline, three tasks are crafted from LongBench-MultiNews dataset for English news and three tasks from the LongBench-LSHT dataset for Chinese news \citep{longbench}, by random sampling of 10, 20 and 30 topics with the context reserved and an index added at the front of each passage. OLP with other non-OLP techniques are tested on the six tasks with four key aspects in each topic, namely “Facts”, “Opinions”, “Assumptions” and “Future plans”. Both OLP and non-OLP outputs are compared with the right answer and the recall rates are calculated by the passage similarity under each topic. As shown in Table \ref{tab_ablation}, the recall rates without OLP decrease drastically with increasing number of topics, whereas the recall rates with OLP range stably between 84.9\% and 93.6\%, which are much higher than those without OLP pipeline with an average increase of 53.6 percentage points.

\section{Conclusion}

In this study, OLP pipeline is proposed for online long-context processing with a collective of well-defined roles namely Topic finder, Topic locator, Relationship checker, Content organizer, Format checker, and Chunk splitter, which cooperate closely to organize the streaming transcript into the topics and aspects that we focus on. Role-RL framework with different update strategies is also developed for better OLP performance by assigning distinct LLMs to their optimal roles according to the judgments from board member LLMs. A benchmark is achieved on our newly compiled live commerce OLP-MINI dataset with an average recall rate of 93.2\% and an LLM cost reduction of 79.4\%, and the effectiveness of our pipeline is also demonstrated on the LongBench dataset which is restructured to OLP format. It is believed by the authors that the applications of OLP pipeline and Role-RL framework are not limited to live transcripts and their great potential awaits further exploration and development in future works.


\begin{thebibliography}{28}
\providecommand{\natexlab}[1]{#1}

\bibitem[{Abbasiantaeb et~al.(2024)Abbasiantaeb, Yuan, Kanoulas, and Aliannejadi}]{ab1}
Abbasiantaeb, Z.; Yuan, Y.; Kanoulas, E.; and Aliannejadi, M. 2024.
\newblock Let the llms talk: Simulating human-to-human conversational qa via zero-shot llm-to-llm interactions.
\newblock In \emph{Proceedings of the 17th ACM International Conference on Web Search and Data Mining}, 8--17.

\bibitem[{Bai et~al.(2023)Bai, Lv, Zhang, Lyu, Tang, Huang, Du, Liu, Zeng, Hou et~al.}]{longbench}
Bai, Y.; Lv, X.; Zhang, J.; Lyu, H.; Tang, J.; Huang, Z.; Du, Z.; Liu, X.; Zeng, A.; Hou, L.; et~al. 2023.
\newblock Longbench: A bilingual, multitask benchmark for long context understanding.
\newblock \emph{arXiv preprint arXiv:2308.14508}.

\bibitem[{Deshpande et~al.(2020)Deshpande, Rajput, Pullapalli, Alluri, Shetty, and Iyer}]{md}
Deshpande, A.; Rajput, A.; Pullapalli, S.; Alluri, S.; Shetty, S.; and Iyer, S. 2020.
\newblock Study of impact of online streaming services (OSS) on youth of 18 to 24 years group with reference to navi mumbai.
\newblock \emph{International Journal of Multi Dimensional Research}.

\bibitem[{Ding et~al.(2024)Ding, Zhang, Zhang, Xu, Shang, Xu, Yang, and Yang}]{long3}
Ding, Y.; Zhang, L.~L.; Zhang, C.; Xu, Y.; Shang, N.; Xu, J.; Yang, F.; and Yang, M. 2024.
\newblock Longrope: Extending llm context window beyond 2 million tokens.
\newblock \emph{arXiv preprint arXiv:2402.13753}.

\bibitem[{Falkowski-Gilski and Uhl(2020)}]{md2}
Falkowski-Gilski, P.; and Uhl, T. 2020.
\newblock Current trends in consumption of multimedia content using online streaming platforms: A user-centric survey.
\newblock \emph{Computer Science Review}, 37: 100268.

\bibitem[{Fan et~al.(2024)Fan, Liu, Liu, Lo, Xia, and Li}]{ab2}
Fan, L.; Liu, J.; Liu, Z.; Lo, D.; Xia, X.; and Li, S. 2024.
\newblock Exploring the Capabilities of LLMs for Code Change Related Tasks.
\newblock \emph{arXiv preprint arXiv:2407.02824}.

\bibitem[{Guo et~al.(2024)Guo, Huang, Liu, Fan, V{\'e}lez, Wu, Wang, Griffiths, and Wang}]{elect}
Guo, X.; Huang, K.; Liu, J.; Fan, W.; V{\'e}lez, N.; Wu, Q.; Wang, H.; Griffiths, T.~L.; and Wang, M. 2024.
\newblock Embodied llm agents learn to cooperate in organized teams.
\newblock \emph{arXiv preprint arXiv:2403.12482}.

\bibitem[{Jiang et~al.(2024)Jiang, Xu, Zhu, Han, Zhang, and Zhu}]{pe5}
Jiang, G.; Xu, M.; Zhu, S.-C.; Han, W.; Zhang, C.; and Zhu, Y. 2024.
\newblock Evaluating and inducing personality in pre-trained language models.
\newblock \emph{Advances in Neural Information Processing Systems}, 36.

\bibitem[{Jin et~al.(2024)Jin, Han, Yang, Jiang, Liu, Chang, Chen, and Hu}]{long4}
Jin, H.; Han, X.; Yang, J.; Jiang, Z.; Liu, Z.; Chang, C.-Y.; Chen, H.; and Hu, X. 2024.
\newblock Llm maybe longlm: Self-extend llm context window without tuning.
\newblock \emph{arXiv preprint arXiv:2401.01325}.

\bibitem[{Kil et~al.(2024)Kil, Mai, Lee, Wang, Cheng, Wang, Liu, Chowdhury, and Chao}]{ab3}
Kil, J.; Mai, Z.; Lee, J.; Wang, Z.; Cheng, K.; Wang, L.; Liu, Y.; Chowdhury, A.; and Chao, W.-L. 2024.
\newblock CompBench: A Comparative Reasoning Benchmark for Multimodal LLMs.
\newblock \emph{arXiv preprint arXiv:2407.16837}.

\bibitem[{Lee et~al.(2024)Lee, Lim, Han, Oh, Chae, Chung, Kim, Kwak, Lee, Lee et~al.}]{pe1}
Lee, S.; Lim, S.; Han, S.; Oh, G.; Chae, H.; Chung, J.; Kim, M.; Kwak, B.-w.; Lee, Y.; Lee, D.; et~al. 2024.
\newblock Do LLMs Have Distinct and Consistent Personality? TRAIT: Personality Testset designed for LLMs with Psychometrics.
\newblock \emph{arXiv preprint arXiv:2406.14703}.

\bibitem[{Li, Patel, and Du(2023)}]{PRD}
Li, R.; Patel, T.; and Du, X. 2023.
\newblock Prd: Peer rank and discussion improve large language model based evaluations.
\newblock \emph{arXiv preprint arXiv:2307.02762}.

\bibitem[{Li et~al.(2024)Li, He, Guo, Bu, Bai, Liu, Liu, Qu, Li, Ouyang et~al.}]{graphreader}
Li, S.; He, Y.; Guo, H.; Bu, X.; Bai, G.; Liu, J.; Liu, J.; Qu, X.; Li, Y.; Ouyang, W.; et~al. 2024.
\newblock GraphReader: Building Graph-based Agent to Enhance Long-Context Abilities of Large Language Models.
\newblock \emph{arXiv preprint arXiv:2406.14550}.

\bibitem[{Liang et~al.(2023)Liang, He, Jiao, Wang, Wang, Wang, Yang, Tu, and Shi}]{MAD}
Liang, T.; He, Z.; Jiao, W.; Wang, X.; Wang, Y.; Wang, R.; Yang, Y.; Tu, Z.; and Shi, S. 2023.
\newblock Encouraging divergent thinking in large language models through multi-agent debate.
\newblock \emph{arXiv preprint arXiv:2305.19118}.

\bibitem[{Lin et~al.(2024)Lin, Peng, Zhang, Sun, Li, Zhao, Xiao, Xu, Qiu, Li et~al.}]{long1}
Lin, B.; Peng, T.; Zhang, C.; Sun, M.; Li, L.; Zhao, H.; Xiao, W.; Xu, Q.; Qiu, X.; Li, S.; et~al. 2024.
\newblock Infinite-llm: Efficient llm service for long context with distattention and distributed kvcache.
\newblock \emph{arXiv preprint arXiv:2401.02669}.

\bibitem[{Liu et~al.(2024)Liu, Yuan, Hu, Li, Chen, Liu, Lu, and Jia}]{RL-GPT}
Liu, S.; Yuan, H.; Hu, M.; Li, Y.; Chen, Y.; Liu, S.; Lu, Z.; and Jia, J. 2024.
\newblock RL-GPT: Integrating Reinforcement Learning and Code-as-policy.
\newblock \emph{arXiv preprint arXiv:2402.19299}.

\bibitem[{Liu et~al.(2023)Liu, Zhang, Li, Liu, and Yang}]{DyLAN}
Liu, Z.; Zhang, Y.; Li, P.; Liu, Y.; and Yang, D. 2023.
\newblock Dynamic llm-agent network: An llm-agent collaboration framework with agent team optimization.
\newblock \emph{arXiv preprint arXiv:2310.02170}.

\bibitem[{Ma et~al.(2024)Ma, Yang, Xiong, Chen, Yu, Zhang, May, Zettlemoyer, Levy, and Zhou}]{long2}
Ma, X.; Yang, X.; Xiong, W.; Chen, B.; Yu, L.; Zhang, H.; May, J.; Zettlemoyer, L.; Levy, O.; and Zhou, C. 2024.
\newblock Megalodon: Efficient llm pretraining and inference with unlimited context length.
\newblock \emph{arXiv preprint arXiv:2404.08801}.

\bibitem[{Pan et~al.(2024)Pan, Lu, Wang, Zheng, Wen, Feng, Zhu, and Chen}]{agentcoord}
Pan, B.; Lu, J.; Wang, K.; Zheng, L.; Wen, Z.; Feng, Y.; Zhu, M.; and Chen, W. 2024.
\newblock AgentCoord: Visually Exploring Coordination Strategy for LLM-based Multi-Agent Collaboration.
\newblock \emph{arXiv preprint arXiv:2404.11943}.

\bibitem[{Pan and Zeng(2023)}]{pe4}
Pan, K.; and Zeng, Y. 2023.
\newblock Do llms possess a personality? making the mbti test an amazing evaluation for large language models.
\newblock \emph{arXiv preprint arXiv:2307.16180}.

\bibitem[{Serapio-Garc{\'\i}a et~al.(2023)Serapio-Garc{\'\i}a, Safdari, Crepy, Sun, Fitz, Romero, Abdulhai, Faust, and Matari{\'c}}]{pe2}
Serapio-Garc{\'\i}a, G.; Safdari, M.; Crepy, C.; Sun, L.; Fitz, S.; Romero, P.; Abdulhai, M.; Faust, A.; and Matari{\'c}, M. 2023.
\newblock Personality traits in large language models.
\newblock \emph{arXiv preprint arXiv:2307.00184}.

\bibitem[{Shinn et~al.(2024)Shinn, Cassano, Gopinath, Narasimhan, and Yao}]{reflexion}
Shinn, N.; Cassano, F.; Gopinath, A.; Narasimhan, K.; and Yao, S. 2024.
\newblock Reflexion: Language agents with verbal reinforcement learning.
\newblock \emph{Advances in Neural Information Processing Systems}, 36.

\bibitem[{Sun et~al.(2023)Sun, Liu, Wang, Zhu, and Iyyer}]{pearl}
Sun, S.; Liu, Y.; Wang, S.; Zhu, C.; and Iyyer, M. 2023.
\newblock Pearl: Prompting large language models to plan and execute actions over long documents.
\newblock \emph{arXiv preprint arXiv:2305.14564}.

\bibitem[{Talebirad and Nadiri(2023)}]{harness}
Talebirad, Y.; and Nadiri, A. 2023.
\newblock Multi-agent collaboration: Harnessing the power of intelligent llm agents.
\newblock \emph{arXiv preprint arXiv:2306.03314}.

\bibitem[{Wei et~al.(2022)Wei, Wang, Schuurmans, Bosma, Xia, Chi, Le, Zhou et~al.}]{COT}
Wei, J.; Wang, X.; Schuurmans, D.; Bosma, M.; Xia, F.; Chi, E.; Le, Q.~V.; Zhou, D.; et~al. 2022.
\newblock Chain-of-thought prompting elicits reasoning in large language models.
\newblock \emph{Advances in neural information processing systems}, 35: 24824--24837.

\bibitem[{Yuan, Liberman, and Cieri(2006)}]{spk}
Yuan, J.; Liberman, M.; and Cieri, C. 2006.
\newblock Towards an integrated understanding of speaking rate in conversation.
\newblock In \emph{Ninth International Conference on Spoken Language Processing}.

\bibitem[{Zhang et~al.(2024)Zhang, Sun, Chen, Pfister, Zhang, and Arik}]{COA}
Zhang, Y.; Sun, R.; Chen, Y.; Pfister, T.; Zhang, R.; and Arik, S.~{\"O}. 2024.
\newblock Chain of Agents: Large Language Models Collaborating on Long-Context Tasks.
\newblock \emph{arXiv preprint arXiv:2406.02818}.

\bibitem[{Zhao et~al.(2024)Zhao, Zu, Xu, Lu, He, Ding, Gui, Zhang, and Huang}]{longagent}
Zhao, J.; Zu, C.; Xu, H.; Lu, Y.; He, W.; Ding, Y.; Gui, T.; Zhang, Q.; and Huang, X. 2024.
\newblock LongAgent: Scaling Language Models to 128k Context through Multi-Agent Collaboration.
\newblock \emph{arXiv preprint arXiv:2402.11550}.

\end{thebibliography}


\onecolumn
\section{Appendix}\label{app:e}

\subsection{An example of OLP result on the OLP-MINI dataset}\label{app:e1}

\begin{mybox1}
\par
\medskip
    \textbf{Topic: Anchor prime one hundred watts wall charger} \\

    \vspace{0pt}
    
    (1)	Opening:
    
    \begin{mybox0}
    [54461, “Alright guys moving on to our next item here we have another prime, so we have anchor prime the one hundred watt wall charger."]
    \end{mybox0}

(2)	Product Description:

    \begin{mybox0}
    [54462, “The last one was sixty seven watts. This one is one hundred watts, so this one is thirty three watts stronger."]

[54463, “It has three ports again a usb, a two usb c outlets and anchor has done the work for you. As you can see the middle outlet, the usb c."]

[54464, “So that's going to be the outlet you use for your laptop, and then the outlet just below has a phone attached to it, so that one's telling you that's the best outlet to charge your phone, okay."]

[54465, “It's great to charge your phone, your tablet, your notebooks, your laptops, all from one single charger."]

[54466, “And again, experience exceptional power and a remarkably compact charger that is forty three percent smaller than the original charger for the macbook."]
    \end{mybox0}

(3)	Price:

    \begin{mybox0}
    [54468, “That discount code will be anchored live eleven. You're getting twenty five percent off and that brings your price down to six thousand three hundred seventy five."]
    \end{mybox0}

(4)	Order Urging:

    \begin{mybox0}
    [54467, “Ultra sleek, ultra lightweight, something to throw in the bag while you are on the go, so go ahead and throw this one in your cart."]
    \end{mybox0}

\medskip

\end{mybox1}


\subsection{An example of OLP result on the LongBench dataset}\label{app:e2}

\begin{mybox1}
\par
\medskip
    \textbf{Topic: Trump Denies Russian Scandal, Calls Media “Fake News"; CNN Contributor Jeffrey Lord Defends Press Conference Performance} \\

    \vspace{0pt}
    
    (1)	Facts:
    
    \begin{mybox0}
    [21995, “President Donald Trump’s press conference had many people talking. 
    Addressing the scandal involving former national security advisor Gen. Michael Flynn, Trump said ‘Russia is fake news.’”]

    [22000, “Tapper later mentioned a reporter from NBC who called out President Trump for incessantly calling the media ‘fake news’ while also ‘saying things that are not true,’ noting the reporter was asking, ‘How do you have credibility to call us fake news?’”]
    \end{mybox0}

(2)	Opinions:

    \begin{mybox0}
    [21996, “In reality, Trump gave a weapons-grade crazy 77-minute press conference. However, CNN contributor and Trump supporter Jeffrey Lord had a different take on that ‘meltdown.’”]

[21997, “‘From my perspective, I thought he was relaxed, he was funny, he was on point. He took the whole issue of the media, and he had a very candid conversation,’ Lord said. ‘This is the kind of conversation that I personally had with him a couple years ago in which he said some version of this same thing that he said today, except he was more specific.’”]

[21998, “Lord described the president as ‘candid, very dedicated to the job, very in command.’”]

[22001, “‘I think it was a simple mistake,’ replied Lord, calling the reporter’s question ‘nitpicky.’ He then proceeded to launch an attack on the network.”]

    \end{mybox0}

(3)	Assumptions:

    \begin{mybox0}

    [21999, “‘What about some of the back and forth that he had with reporters? Is that really presidential four weeks into a new term?’ Blitzer asked. ‘I think this is going to be his presidential style,’ Lord replied.”]
    \end{mybox0}

(4)	Future Plans:

    \begin{mybox0}
    [None in the provided text]
    \end{mybox0}
    
\medskip
\end{mybox1}

\end{document}